\documentclass[conference]{IEEEtran}
\IEEEoverridecommandlockouts
\usepackage{cite}
\usepackage{amsmath,amssymb,amsfonts}
\usepackage{algorithmic}
\usepackage{graphicx}
\usepackage{textcomp}
\usepackage{algorithm}
\usepackage{colortbl}
\usepackage{multirow}
\usepackage{soul}
\usepackage[dvipsnames]{xcolor}

\def\BibTeX{{\rm B\kern-.05em{\sc i\kern-.025em b}\kern-.08em
    T\kern-.1667em\lower.7ex\hbox{E}\kern-.125emX}}
\begin{document}

\title{Incorporating Feature Pyramid Tokenization and Open Vocabulary Semantic Segmentation}

\author{\IEEEauthorblockN{Jianyu Zhang, Li Zhang, Shijian Li}
\IEEEauthorblockA{\textit{Zhejiang University}\\
Hangzhou, China \\
\{jianyu.zhang,zhangli85,shijianli\}@zju.edu.cn}
}

\maketitle

\begin{abstract}
  The visual understanding are often approached from 3 granular levels: image, patch and pixel. Visual Tokenization, trained by self-supervised reconstructive learning, compresses visual data by codebook in patch-level with marginal information loss, but the visual tokens does not have semantic meaning. Open Vocabulary semantic segmentation benefits from the evolving Vision-Language models (VLMs) with strong image zero-shot capability, but transferring image-level to pixel-level understanding remains an imminent challenge. 
  In this paper, we treat segmentation as tokenizing pixels and study a united perceptual and semantic token compression for all granular understanding and consequently facilitate open vocabulary semantic segmentation. 
  Referring to the cognitive process of pretrained VLM where the low-level features are progressively composed to high-level semantics, we propose Feature Pyramid Tokenization (PAT) to cluster and represent multi-resolution feature by learnable codebooks and then decode them by joint learning pixel reconstruction and semantic segmentation. 
  We design loosely coupled pixel and semantic learning branches. The pixel branch simulates bottom-up composition and top-down visualization of codebook tokens, while the semantic branch collectively fuse hierarchical codebooks as auxiliary segmentation guidance.
  Our experiments show that PAT enhances the semantic intuition of VLM feature pyramid, improves performance over the baseline segmentation model and achieves competitive performance on open vocabulary semantic segmentation benchmark. Our model is parameter-efficient for VLM integration and flexible for the independent tokenization. We hope to give inspiration not only on improving segmentation but also on semantic visual token utilization. 
\end{abstract}

\section{Introduction}

Open Vocabulary Semantic Segmentation (\textbf{OVSS}) aims at recognizing pixels by open set categories. Benefit from powerful zero-shot vision-language model (VLM) epitomized by CLIP \cite{CLIP}, the research have reduced the challenge to transferring image-level understanding to pixel-level. 
In general, existing OVSS methods have explored following strategies: transferring CLIP representation by spatial annotations \cite{OpenSeg,MaskCLIP,SimSeg,OVSeg}, aligning with spatial-aware models like SAM and Diffusion \cite{SAM,OVSAM,PosSAM,OVDiff,ODISE}, denoising dense representation of VLM \cite{CAT-Seg,CLIPSelf,CLIP-DINOiser}. Despite the merits, these methods either require heavy training or overlook the potential of intermediate pretrained features. 
Prior studies \cite{ViT-Learn,SaliencyMaps,UnderstandCNN} have revealed layer-wise feature extraction preference from low-level patterns to high-level semantics in the pretrained model, which is an intuitive process of visual concept formulation. Some traditional semantic segmentation methods \cite{Mask2Former,kMax,MSMFormer} benefit from multi-stage feature integration through Feature Pyramid Network (FPN) \cite{FPN} and U-Net \cite{U-Net}. Therefore, unlocking potential of pretrained VLM feature pyramid is crucial for open vocabulary segmentation task and has been proven successful in existing works \cite{SAN,FC-CLIP,SED}. From another perspective, segmentation can be viewed as tokenizing pixel by semantic labels. Traditional visual tokenization (i.e. VQVAE series \cite{VQVAE,VQGAN,ViT-VQGAN}) learns a perceptual, non-semantic compression of image for generation task. Applications \cite{SD,BEiT3} of visual tokenization utilize its effective perceptual compression to accelerate diffusion or multi-modal interaction, but semantic visual tokenization is barely explored.

\begin{figure}[!t]
	\centering
	\includegraphics[trim=60 120 60 100, width=0.9\linewidth]{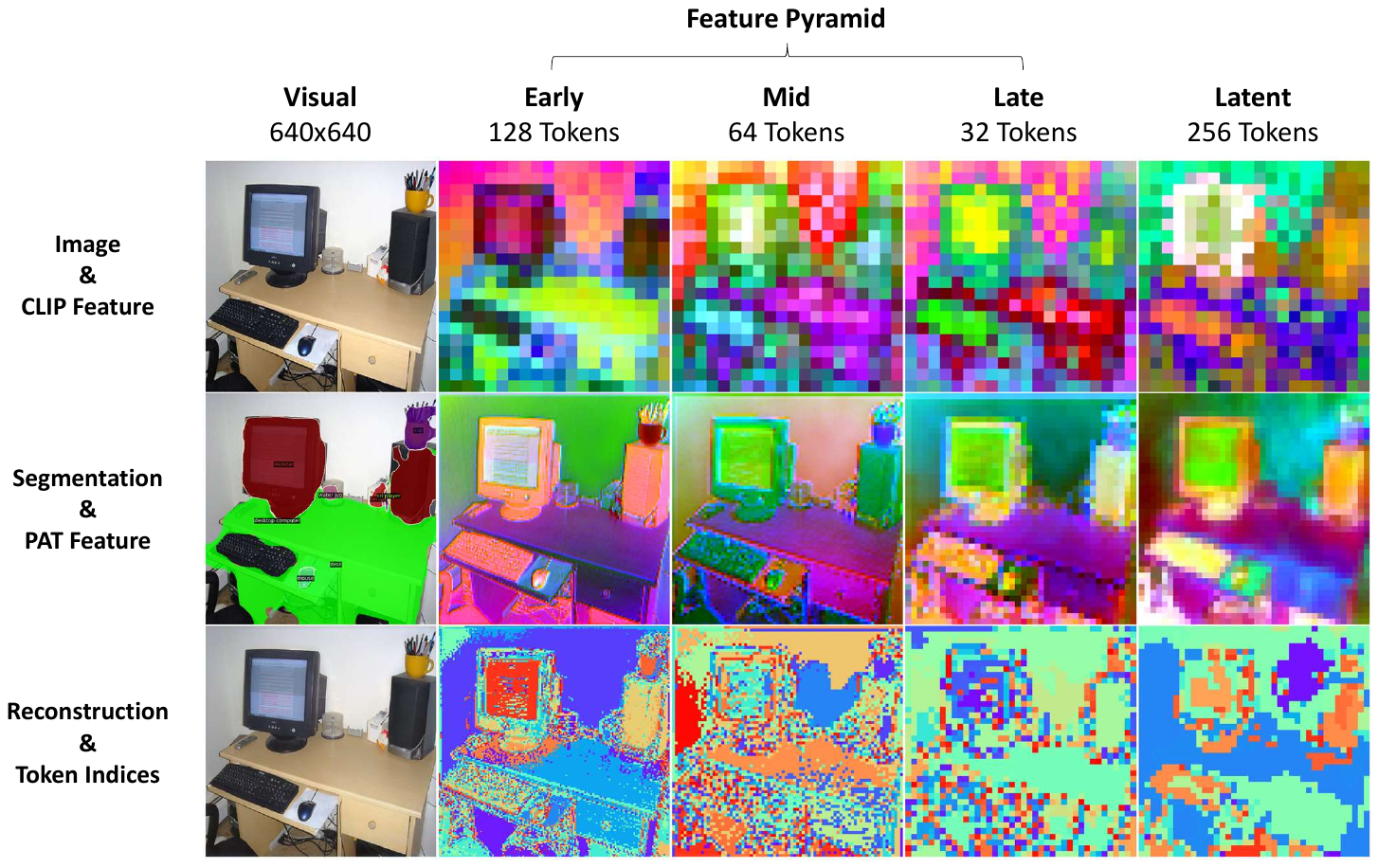}
    \caption{Open Vocabulary Segmentation example. After learning pyramid tokenization by segmentation and reconstruction, the tokenized feature pyramid (row 2) demonstrates vastly improved semantic intuition than the original pretrained representation (row 1) at each stage. The semantic concepts compose from \textbf{Early} low-level color, edge to \textbf{Mid/Late} parts, structures, textures and then finally the \textbf{Latent} objects. At row 3, PAT shows feature clustered into meta semantic tokens which are comprehensive and disentangled.  }
	\label{fig:compare}
\end{figure}

In this paper, we explore a heuristic meta semantic discovery by combining visual tokenization with segmentation. We propose \textbf{PAT}: feature pyramid tokenization to extract multi-resolution visual tokens as intermediate segmentation. The visual tokens connect the image- and pixel-level understanding by different granular semantic. PAT is learned by reconstruction and segmentation using a lightweight shared decoder while keeping tokenization detachable. Our method aligns the cognitive process of human and pretrained visual representation model, where low-level visual patterns coalesce into abstract concepts by sensation, and concepts are progressively refined with details by visualization.

We benchmark our method on common OVSS evaluation protocol, where PAT is trained on COCO Stuff and tested on Pascal Context-59 , Pascal Context-459, ADE20K-150, and ADE20K-847 \cite{COCO-Stuff,PContext,ADE20k}. 
PAT showcases efficacy in extracting semantic-rich visual token pyramid, the improvements to the baselines, and the competitive performance to the state-of-the-art (SOTA) methods. We hope PAT can prompt rethinking of prevailing paradigms in pretrained representation utilization.

\begin{figure}[t]
	\centering
	\includegraphics[trim=50 160 50 90, width=.9\linewidth]{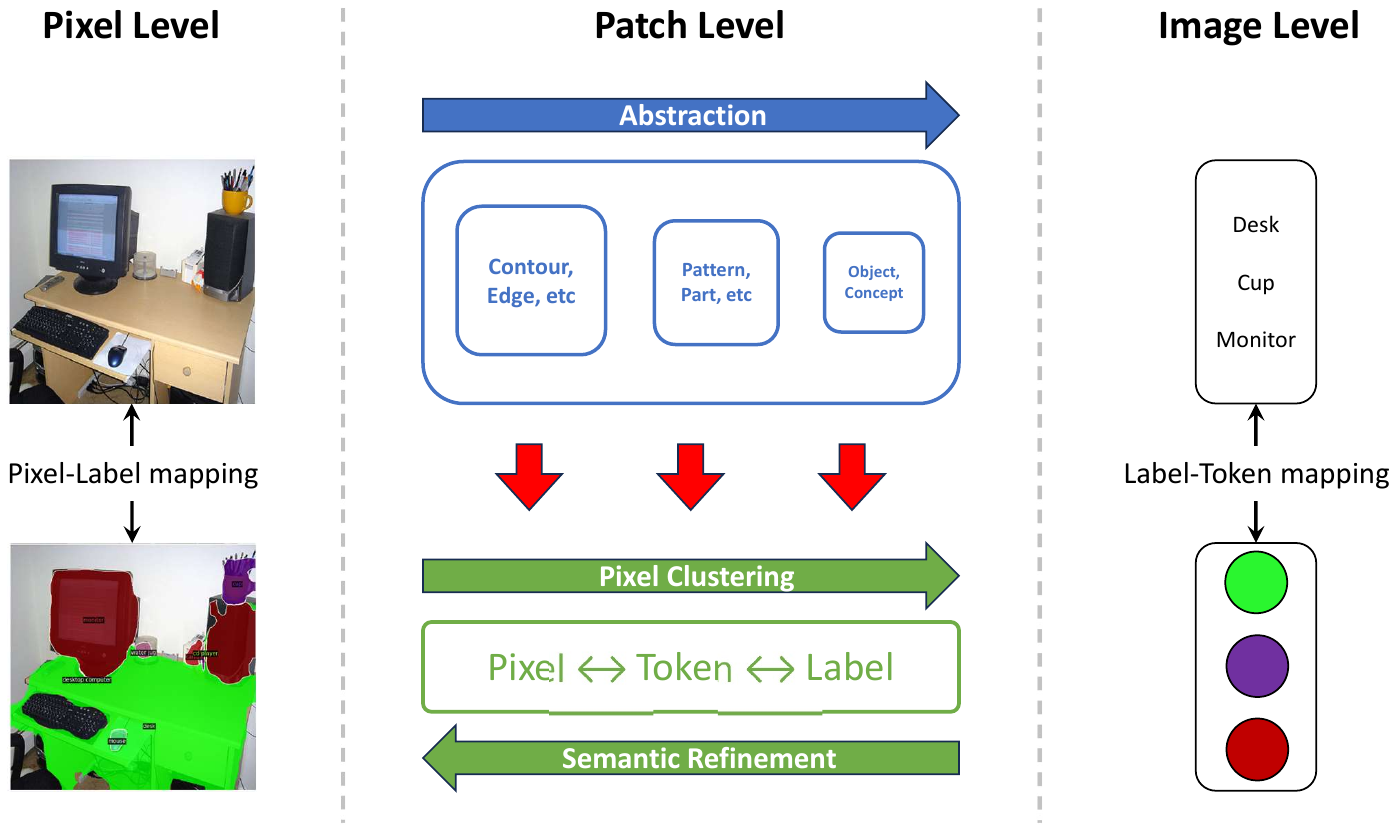}
    \caption{PAT concept. The image-level and pixel-level understanding gap brings difficulty to use VLM for segmentation (i.e. \textit{Pixel-Label} mapping). PAT clusters and tokenizes feature pyramid to abstract \textit{Tokens}, then refines semantic progressively to pixel-level based on the easily learned \textit{Label-Token} mapping. The \textit{Tokens} become the patch-level bridge between \textit{Labels} and \textit{Pixels}.}
	\label{fig:overview}
\end{figure}

\section{Related Work}

\subsection{Visual Tokenization}
Visual tokenization is regarded as a compression of visual data. At first VQVAE \cite{VQVAE} tokenize continuous representations by a finite set of learnable tokens through image reconstruction. The follow-up works advance by generative adversarial model \cite{VQGAN} and Transformer \cite{ViT-VQGAN}. 
The pretrained visual tokens \cite{VQGAN} serve as input for VLMs \cite{BEiT3,PeCo} by its language-alike form. However, as \cite{SD} pointed out, autoregressive tokens compress perception instead of semantic. 
The multi-modal Large Language Model (LLM) further elevates importance of visual semantic compression. \cite{BLIP2} compresses frozen image encoder to few tokens as LLM input for efficient multi-modal interactions. \cite{SPAE} instead uses frozen LLM vocabulary and VLM guidance to create hierarchical language-aware visual tokenization. 
PAT explores tokenizing pretrained VLM feature directly and obtain semantic-rich tokens at low cost.

\subsection{Open Vocabulary Semantic Segmentation}
Open Vocabulary Semantic Segmentation is challenging to learn per-pixel semantic from open-set comprehensive visual concepts. Due to the expensive per-pixel annotation, research tend to finetune accessible pretrained representations. \cite{MaskCLIP,OpenSeg,ZSSeg} align VLMs with spatial regions to obtain semantic locality. \cite{OVSAM,PosSAM} integrate SAM \cite{SAM} for unlimited pixel annotation. 
Despite success for finetune-segment paradigm, the compute and data efficiency are also critical for real world application. One-stage end-to-end methods with closer VLM integrations \cite{CAT-Seg,SAN,FC-CLIP,SED} are proposed for faster inference speed and better performances. 
In addition to segmentation focused designs, the intermediate representation quality is also vital for transferring image understanding to pixel-level, which is approached by (1) aligning with perceptual Diffusion latent \cite{OVDiff,ODISE} can be beneficial for segmentation; (2) refining VLM training \cite{CLIPSelf,CLIP-DINOiser} and (3) denoising the pretrained VLM features \cite{DVT,FeatUp}. Trading off the computational cost, we explore tokenizing pretrained VLM features to provide perceptually compressed tokens for open vocabulary supplementary. 

Based on above, we try to \textit{unify pixel and semantic encoding} by assuming each codebook represents a set of meta semantic, a mixture of perceptual and comprehensive visual concept, and the tokenizing feature becomes code-based segmentation. From this perspective, the codebook may serve both perceptual compression and semantic segmentation. 
To verify our assumption, we visualize pretrained VLM feature in Fig\ref{fig:compare} row 1. The cognitive mechanism \cite{ViT-Learn} in pretrained feature pyramid is partially observable, where different VLM stage held different feature extraction preference roughly divided to low-level contours, mid-level structures and high-level concepts (see supplementary for more examples). However, the patch-level visual concepts are noisy and lack of details for segmentation.  
Therefore, as specified in Fig\ref{fig:overview}, we leverage pretrained feature pyramid tokenization carrying multi-granular semantics as a bridge between pixels and image-level semantics. During abstraction, pixel-level features are sequentially denoised and clustered into patch-level meta semantic tokens. For refinement, label associated tokens guide the decoding of the meta semantic tokens to both pixels and dense labels.

\section{Method}

\begin{figure*}[!htb]
	\centering
	\includegraphics[trim=30 130 30 130, width=0.8\linewidth]{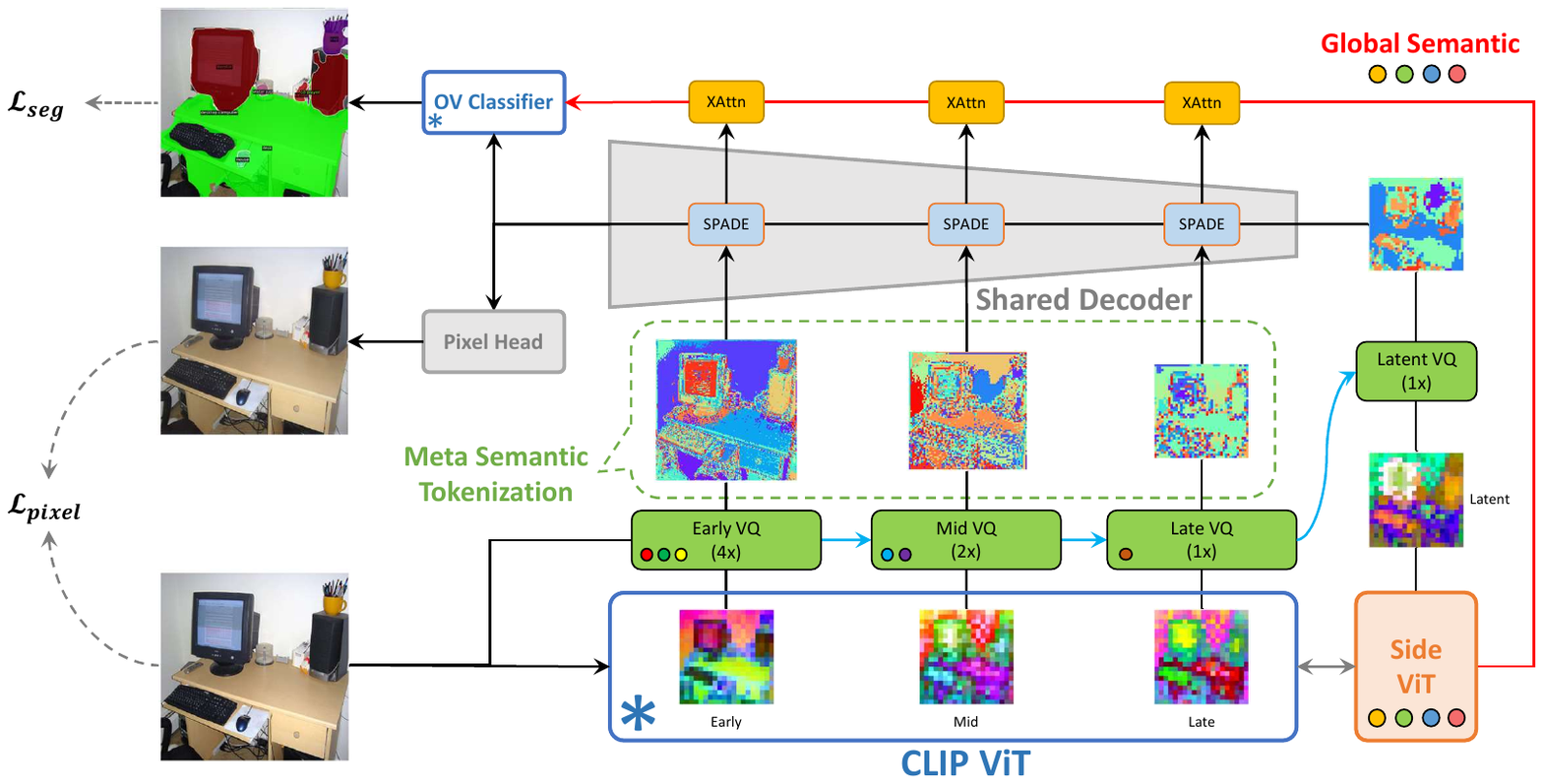}
    \caption{PAT architecture. The global tokens from \textcolor{orange}{Side ViT} focus on CLIP-aware semantic concepts and local tokens (in \textcolor{YellowGreen}{PAT VQ modules}) learn cluster oriented meta semantic \cite{MSMFormer}. The decoupled global and local tokens are mutually guided by reconstruction and segmentation as described in Fig \ref{fig:module}.}
	\label{fig:arch}
\end{figure*}

In this section, we follow 2 previous discussed principals to design our model detailed in Fig\ref{fig:arch}: (1) \textit{tokenization is segmentation} (2) \textit{semantics are composed from pixel detail to abstract concept}. We propose feature Pyramid Tokenization (\textbf{PAT}) to first extracts local feature by multi-stage codebooks to meta semantic, and then uses decoupled token fusion to integrate them to a side adapter segmentation model. Finally, the local tokens are decoded by a shared decoder unifying pixel and semantic learning. By end-to-end training on COCO-Stuff \cite{COCO-Stuff}, PAT extracts clean and semantic-rich tokens (Fig\ref{fig:compare}) and improves segmentation performance competitive to SOTA. 

\label{sec:cluster}
\subsection{Feature Pyramid Tokenization}

As Fig\ref{fig:overview} demonstrated, the pretrained VLM feature pyramid has embedded with semantic information from detail to abstract, and tokenization creates meta semantics to link the pixel and label. However, directly tokenizing pretrained feature (Fig\ref{fig:compare} row 1) cannot produce desirable segmentation due to the observable noise and relatively low resolution. To denoise the pretrained feature for cleaner tokenization and elevates the detail-to-abstract cognition, we propose to \textit{clustering on multi-resolution feature pyramid}. 
We first claim that tokenization and feature clustering are special cross attention and can be geometrically aligned by Eq\ref{eq:hs-attn}. For simplification, we use \textbf{QKV} form for all module inputs although \textbf{K} and \textbf{V} is practically the same for these modules. $g(\cdot)$ is vector normalization, $\mathsf{scatter}_{\textbf{Q}}(\cdot, \textbf{V})$ indicates feature assigning:
\begin{equation} \label{eq:hs-attn}
\begin{split}
    \mathsf{VQ}(\textbf{Q}, \textbf{K}, \textbf{V})
    &=\mathsf{scatter}_{\textbf{Q}}(\arg\max({\textbf{Q}\textbf{K}^T}),\textbf{V}) \\
    \mathsf{HSAttn}(\textbf{Q}, \textbf{K}, \textbf{V})
    &=g(\mathsf{softmax}(\kappa g(\textbf{Q})g(\textbf{K})^T)\textbf{V}) \\
    \mathsf{Attn}(\textbf{Q}, \textbf{K}, \textbf{V})
    &=\mathsf{softmax}(\textbf{Q}\textbf{K}^T/\sqrt{d_k})\textbf{V} 
\end{split}
\end{equation}

Specifically, $\mathsf{HSAttn}$ \cite{MSMFormer} learns vMF meanshift clustering of \textbf{V} and update centroid \textbf{Q}. The $\mathsf{VQ}$ tokenization can also be expressed by one-hot hard assignment attention from \textbf{V} to \textbf{Q}. The $\mathsf{HSAttn}$ combined with $\mathsf{VQ}$ is essentially using the centroid feature (i.e. codebook token) to represent the corresponding feature cluster, which can be regarded as \textit{meta semantic abstraction}. We provide more visual analysis (Fig\ref{fig:meta}) of the feature clustering and meta semantics on Section \ref{sec:meta}.

The cluster oriented tokenization is implemented in PAT VQ module (Fig\ref{fig:module}). Adopting SAN \cite{SAN} as segmentation backbone, the feature pyramid are split to early, mid and late stages where CLIP features (\textcolor{NavyBlue}{$f_\mathsf{CLIP}$}) fuse with Side Features (\textcolor{orange}{$z$}). The stage-wise codebook clusters and tokenizes the CLIP features into multi-resolution meta semantic pyramid as a bridge representation mixing perception and semantic. With bottom-up codebook fusion, the series of PAT VQ modules simulates meta semantic composition as the current stage meta semantics are conditioned on previous stage.

\label{sec:decouple}
\subsection{Decoupled Pixel and Semantic Branch}

\begin{figure}[!ht]
	\centering
	\includegraphics[trim=40 120 40 100, width=0.9\linewidth]{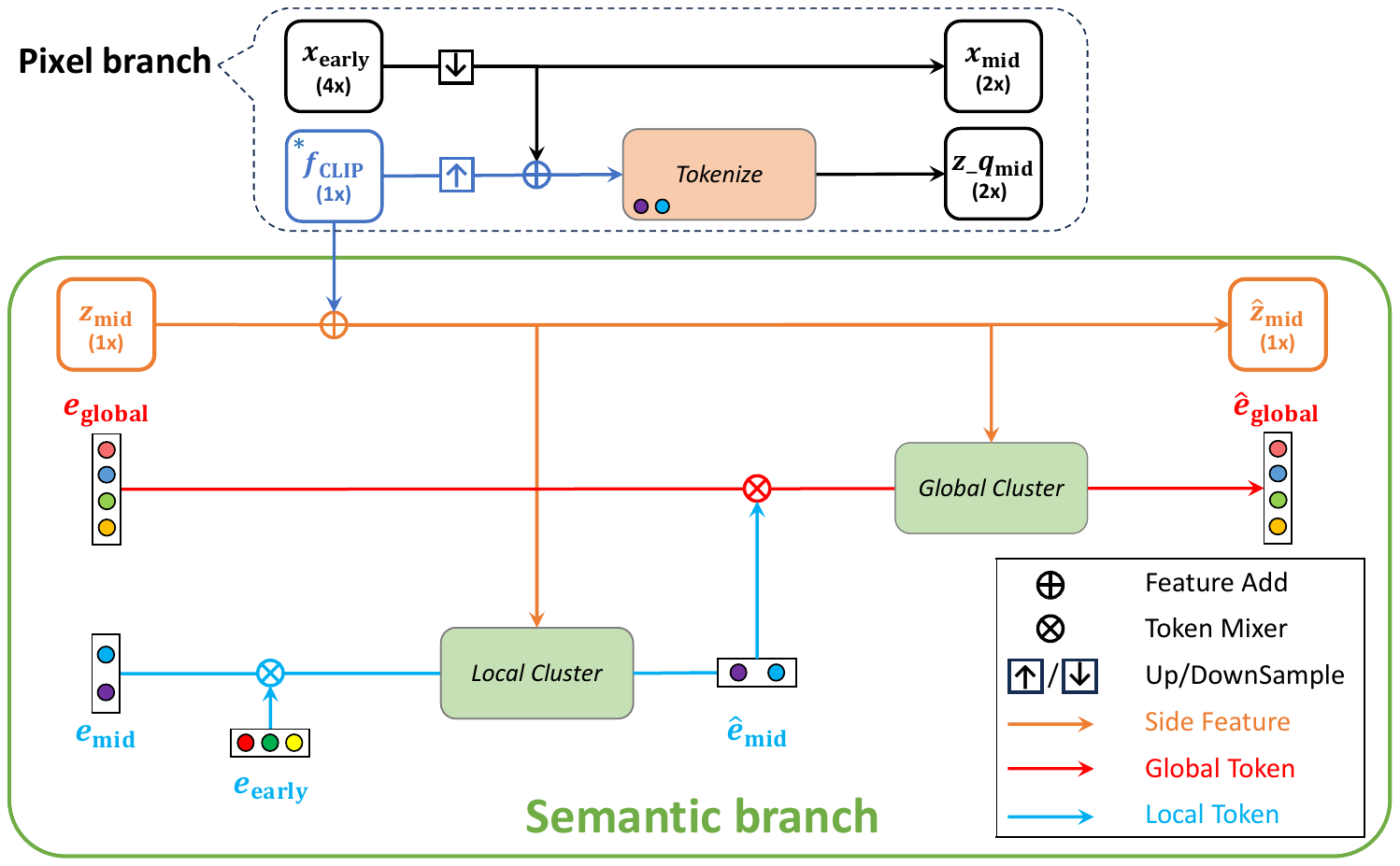}
    \caption{Mid stage PAT VQ module. The module decouples the semantic learning (\textcolor{red}{\textbf{Global Tokens}} with \textcolor{cyan}{\textbf{Local Tokens}} and \textcolor{orange}{\textbf{Side Features}}) and pixel decoding (\textcolor{cyan}{\textbf{Local Tokens}} with \textcolor{NavyBlue}{\textbf{CLIP Features}} and \textbf{Pixel Residual}). }
	\label{fig:module}
\end{figure}

The contradictory between perceptual and semantic compression \cite{SD} brings challenge for aligning meta semantic to actual semantic. PAT abides decoupled design for the local and global tokens to avoid such interference, where the interactions between pixel branch (Fig\ref{fig:module}, top) and semantic branch (Fig\ref{fig:module}, bottom) are indirect. 
Formally, for mid stage, the PAT VQ module has 5 inputs: CLIP features $f_\mathsf{CLIP}$, side features $z_{mid}$, global tokens $e_{global}$, previous stage updated local tokens $\hat{e}_{early}$ and previous stage pixel residual $x_{early}$. After processing, it transmits the updated local tokens $\hat{e}_{mid}$ and downsized pixel residual $x_{mid}$ to the next stage, meanwhile it produces VQ features $z\_q_{mid}$ for the shared Decoder. The updated global tokens $\hat{e}_{global}$ are forwarding in the Side ViT straight to the next stage. The detailed workflow is as follows:
\begin{equation} \label{eq:heratvq}
\begin{split}
    & \mathbf{Pixel\ branch:} \\
    & \quad x_{mid} = \mathsf{DownSampler}(x_{early})  \\
    & \quad z\_q_{mid} = \mathsf{vMFVQ}(e_{mid}, \mathsf{UpSampler}(f_\mathsf{CLIP} + x_{mid}, 2)) \\
    & \mathbf{Semantic\ branch:} \\
    & \quad \hat{z}_{mid} = z_{mid} + f_\mathsf{CLIP} \\
    & \quad \hat{e}_{mid} = \mathsf{HSAttn}(e_{mid} + \mathsf{TokenMixer}(\hat{e}_{early}), z_{mid}, z_{mid}) \\
    & \quad \hat{e}_{global} = \mathsf{Attn}(e_{global} + \mathsf{TokenMixer}(\hat{e}_{mid}), z_{mid}, z_{mid}) \\
\end{split}
\end{equation}

The $\mathsf{vMFVQ}$ is $\mathsf{VQ}$ with vMF prior. The $\mathsf{TokenMixer}$ is 3-layer MLP. The $\mathsf{Up/DownSampler}$ is DySample \cite{DySample} and convolution for scaling features. To further denoise feature, we align $z\_q_{mid}$ to input image $x$ by CRF and TV losses \cite{FeatUp} (the Spatial Alignment Loss group). Since the forward pass has no information flow from global to local tokens, the pixel branch can be separated as independent tokenizer. 

\label{sec:shared}
\subsection{Shared Decoder}

Despite decoupled design, decoding pixel and pixel-level semantic label can share same latent space like in layout-to-image generation \cite{UnlockPTM}. Therefore, we propose a shared lightweight Transformer decoder which adopt U-Net transmission of the stage-wise tokenization $z\_q_{mid}$ and apply SPADE \cite{SPADE} to fuse them to the pixel latent $z_{latent}$ (the VQ output of Side ViT) from top-down. 
Apart from the pixel decoding, the global tokens take additional guidance from the decoded features to enhance multi-resolution semantic fusion. The "mid" stage decoding can be summarized as follows:
\begin{equation} \label{eq:decoder}
\begin{split}
    & \hat{z}_{latent} = \mathsf{SPADE}(\mathsf{UpSampler}(z_{latent}, 2), z\_q_{mid}) \\
    & \hat{e}_{global} = \mathsf{Attn}(e_{global}, \hat{z}_{latent}, \hat{z}_{latent}) \\
\end{split}
\end{equation}

After decoding all tokenization stages, the final $\hat{z}_{latent}$ is eligible for both reconstruction and segmentation. For mask generation, we use product of $\hat{z}_{latent}$ and $\hat{e}_{global}$ after projected into the same space by MLPs. The per-query masks are adopted by CLIP-based open vocabulary classifier to calculate segmentation losses.
For pixel decoding, we simply use a transposed convolution head to recover original image and trained by L1, L2 and perceptual loss \cite{PerceptualLoss} for the Reconstructive Loss group with weight combination specified in \cite{ViT-VQGAN}. 

\section{Experiments}

\begin{table} 
    \centering
    \caption{4 datasets mIoU performance for OVSS models with various VLM backbone. PAT shows competitive performance to SOTAs.}
    \setlength\tabcolsep{2.5pt}
    \renewcommand{\arraystretch}{1.1}
    \scalebox{1.}{\begin{tabular}{l | c | c | r r r r}
        \hline
        Method & VLM & TrainSet & PC59 & A150 & PC459 & A847 \\
        \hline
        OpenSeg \cite{OpenSeg}   & ALIGN & COCO-Panoptic & 40.1   & 17.5  & 7.9 & 4.4     \\
        OVSeg \cite{OVSeg}     & CLIP & COCO-Stuff\&Cap. & 53.3   & 24.8  & 11.0 & 7.1     \\
        CAT-Seg \cite{CAT-Seg} & CLIP & COCO-Stuff & 57.5 & 27.2 & 16.6 & 8.4     \\
        SAN \cite{SAN}     & CLIP & COCO-Stuff & 53.8   & 27.5  & 16.2 & 10.1     \\
        SAN \cite{SAN}     & EVACLIP & COCO-Stuff & 55.1   & 30.4  & 17.7 & 11.0     \\
        SED \cite{SED}     & C. CLIP & COCO-Stuff & 57.3   & 31.6  & 18.6 & 11.4     \\
        CAT-Seg \cite{CAT-Seg} & ft. CLIP & COCO-Stuff & 57.5 & \textbf{31.8} & \textbf{19.0} & 12.0     \\
        \hline
        \rowcolor[gray]{.9} PAT (OURS) & CLIP & COCO-Stuff & 55.1 & 28.4 & 16.8 & 10.4     \\
        \rowcolor[gray]{.9} PAT (OURS) & EVACLIP & COCO-Stuff & \textbf{57.9} & 31.6 & 18.9 & \textbf{12.1}     \\
        \hline
        \hline
        
    \end{tabular}} 
    \label{tab:main-results}
\end{table}

\begin{figure} 
    \centering
    \includegraphics[trim=0 25 0 20, width=0.8\linewidth]{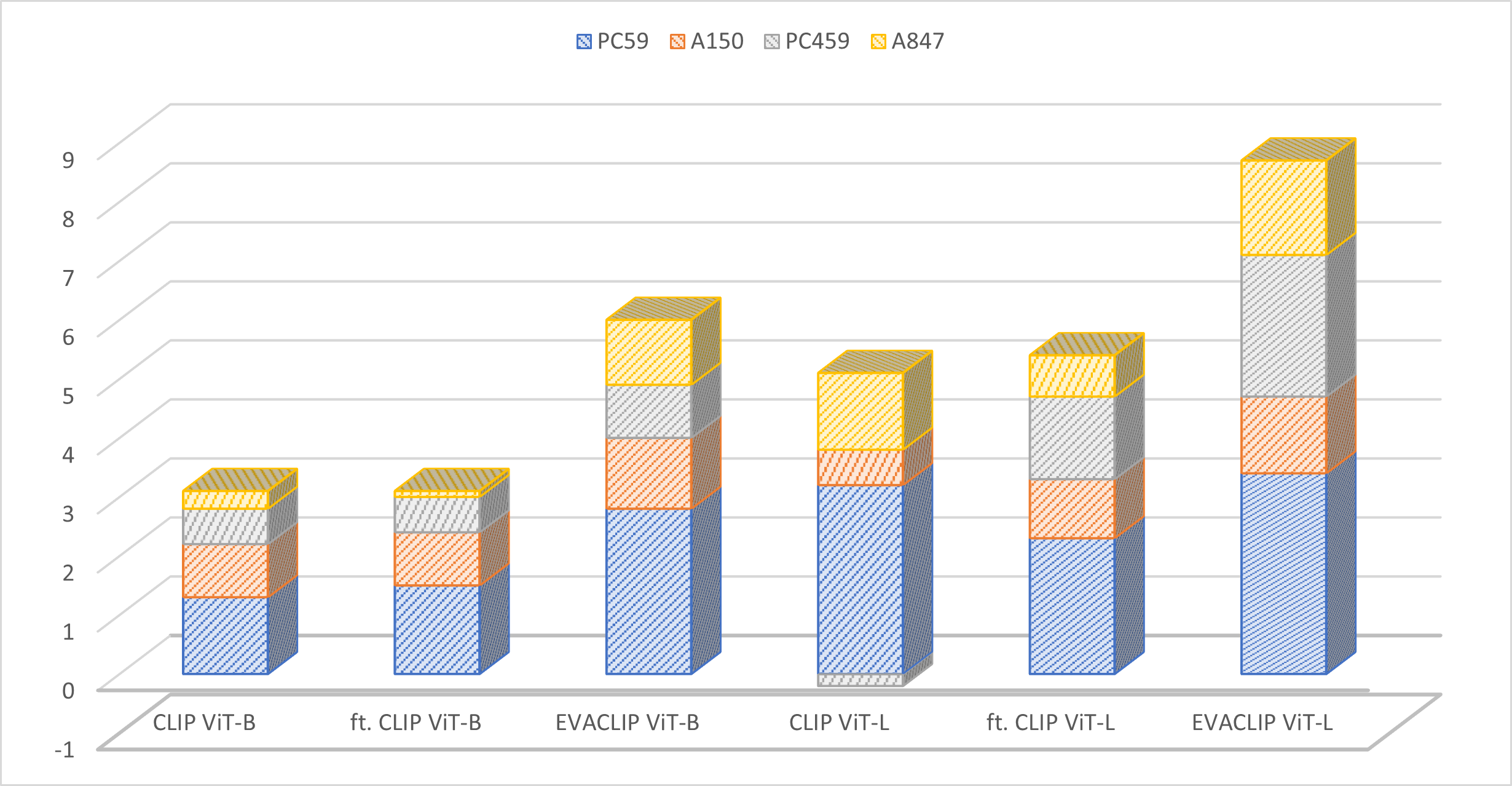}
    \caption{Accumulated SAN baseline improvements using different VLM. }
    \label{fig:improve}
\end{figure}

We follow common OVSS setting to train PAT on COCO-Stuff \cite{COCO-Stuff} and evaluate on 4 datasets \cite{PContext,ADE20k}: Pascal Context-59, Pascal Context-459, ADE20k-150, ADE20k-847. 
We use the class-wise mean intersection over union (mIoU) as metrics and smooth the variance by 5-run average. 
We compare to notable models using pretrained VLMs \cite{OpenSeg,OVSeg,CAT-Seg,SAN,SED} by ensuring mostly fair-comparison under same VLM size setting (e.g. ViT-B/16) and training data distribution, although some of them require different guidance modes like COCO-Captions \cite{COCO-Captions} and COCO-Panoptic \cite{MSCOCO}. SAN \cite{SAN} as the semantic branch is our baseline.

\subsection{Implementation Details}
Our VLM integration is identical to SAN, where Side ViT (240-dim, 8 layers) fuses VLM at \{0, 1/4, 1/2, 3/4\} depth. We treat last 3 fusion depths as the feature pyramid, upscale their resolution by \{4x, 2x, 1x\}. The 3 stage and latent features are tokenized by 32-dim codebook of size \{128, 64, 32, 256\}. The image is augmented and scaled the same as SAN to enable $640^2$ standard resolution. For lightweight Decoder, we use 3 256-dim Transformer+SPADE to fuse tokenized feature pyramid followed by 2 extra Transformer layers. The segmentation loss compilation is same as SAN. 
We use weights 0.1, 0.1, 1.0 for the VQ loss, spatial alignment loss, and reconstructive loss group respectively. 
Our integration adds 23M trainable parameters to the pretrained VLM.
We train PAT with 3 NVIDIA RTX-3090, 12 batch size, 120K iterations, 1e-4 learning rate and 1e-4 weight decay with AdamW optimizer. 

\subsection{Segmentation Results}

Table \ref{tab:main-results} presents PAT comparison to OVSS models. PAT with EVACLIP \cite{EVA-CLIP} demonstrates competitive performances compared to the current state-of-the-art (SOTA). Specifically, the PAT outperforms SED by \textbf{+0.55} mIoU (averaged on 4 datasets) and is comparable to CAT-Seg. The PAT also improves baseline SAN by \textbf{+0.78} and \textbf{+1.6} for CLIP and EVACLIP respectively. 
To further study the relation between OVSS performance and VLM choice, we benchmark baseline improvements for CLIP, finetuned CLIP and EVACLIP in both ViT-B and -L settings. Fig\ref{fig:improve} shows accumulated mIoU gains on 4 datasets, which prove that the cluster oriented feature pyramid tokenization could be more effective and precise on clearer VLM representations, constraint by VLM architecture. More results are provided in the Supplementary. We leave verifying other CLIP versions for the future work due to their incompatible integration with SAN. 

\subsection{Ablation Study}
We conduct ablations on the key designs in Table \ref{tab:ablation}, where we add pixel evaluation of reconstruction FID \cite{FID}. We concludes the novelty of our design in following aspects:

\subsubsection{The Effectiveness of Clustering} 
The cluster orientation has 3 key designs: vMF meanshift by $\mathsf{vMFVQ}$ and $\mathsf{HSAttn}$, the multi-resolution feature pyramid and the spatial alignment losses. Our ablations show that the vMF prior is the most crucial factor as local feature clustering described in Eq\ref{eq:hs-attn}. We also discover that uniform resolution feature pyramid is undesirable for clustering because 4x introduces excessive pixel information and 1x falls short on bottom-up meta semantic aggregation. In addition, the spatial alignment demonstrates its effect on feature denoising by original pixel patterns, although the performance improvement is minor. 

\subsubsection{The Necessity of Decoupled Learning} 
The decoupled design for pixel and semantic information branches is critical to mitigate learning conflicts. Firstly, the $\mathsf{TokenMixer}$ connects the meta semantics in different stage and fuse them to global tokens, without which the local tokens learn better perceptual compression but have no contribution to the segmentation. Secondly, the pixel residual avoids tokenization directly on frozen feature and reduces the training difficulties and further decouple the pixel and semantic learning, proven effective by performance. Lastly, if we use one unified codebook embedding for both local and global tokens, the contradictory has resurfaced by such direct perceptual and semantic mixture. 

\subsubsection{The mutual beneficiary of shared decoding} 
We empirically find the shared decoding can further enhance the semantic propagation through feature pyramid, showing that the meta semantic effectively serves as a bridge between pixel and label. 
By gradually removing the decoding stages of the feature pyramid, we measure the stage-wise contribution to both pixel and semantic branch, where the pixel decoding is more sensitive to the feature pyramid integrity. 

\begin{table} 
    \centering
    \caption{Ablation Study for PAT EVACLIP ViT-B/16 baseline on Coco-Stuff. The vanilla design is empirically the best. }
    \setlength\tabcolsep{2.5pt}
    \renewcommand{\arraystretch}{1.1}
    \scalebox{1.}{\begin{tabular}{l l | r r r r | r}
        \hline
        Ablations & Settings  & PC59 & A150 & PC459 & A847 & rFID$\downarrow$\\
        \hline
        \multirow{2}{*}{\textbf{Baseline}} & FPN = (Early, Mid, Late) & \multirow{2}{*}{\textbf{57.9}} & \multirow{2}{*}{\textbf{31.6}} & \multirow{2}{*}{\textbf{18.9}} & \multirow{2}{*}{\textbf{12.1}} & \multirow{2}{*}{\textbf{13.8}} \\
        & Scale = (4x, 2x, 1x) &  &  &  &      \\
        \hline
        \multirow{3}{*}{Cluster} 
        & No vMF meanshift & 53.0 & 27.6 & 16.5 & 10.3 & 21.6 \\
        & Scale = (4x, 4x, 4x) & 57.1 & 30.9 & 18.1 & 11.4 & 12.4 \\
        & Scale = (1x, 1x, 1x) & 54.5 & 28.7 & 17.2 & 10.8 & 14.4 \\
        & No Spatial Alignment & 57.1 & 31.0 & 18.7 & 11.7 & 15.2 \\
        \hline
        \multirow{3}{*}{Decouple} & No TokenMixer & 55.4 & 29.6 & 17.5 & 11.0 & 12.9 \\
        & No Pixel Residual & 53.9 & 27.1 & 16.8 & 10.2 & 33.6 \\
        & Unified Pix/Seg Tokens & 47.4 & 26.0 & 15.3 & 9.9 & 52.4 \\ 
        \hline
        \multirow{4}{*}{Decoder} & Separate Decoding & 57.5 & 31.2 & 18.4 & 11.5 & 13.7 \\ 
        & FPN = (\st{Early}, Mid, Late) & 56.6 & 30.8 & 18.5 & 11.4 & 25.1 \\ 
        & FPN = (\st{Early}, \st{Mid}, Late) & 56.2 & 29.1 & 18.0 & 10.7 & 31.2 \\ 
        & FPN = (\st{Early}, \st{Mid}, \st{Late}) & 55.8 & 28.5 & 17.6 & 10.5 & 45.8 \\ 
        \hline
    \end{tabular}} 
    \label{tab:ablation}
\end{table}

\label{sec:meta}
\subsection{Feature Clustering and Meta Semantic Analysis}

\begin{figure} 
    \centering
    \includegraphics[trim=0 120 0 120, width=.95\linewidth]{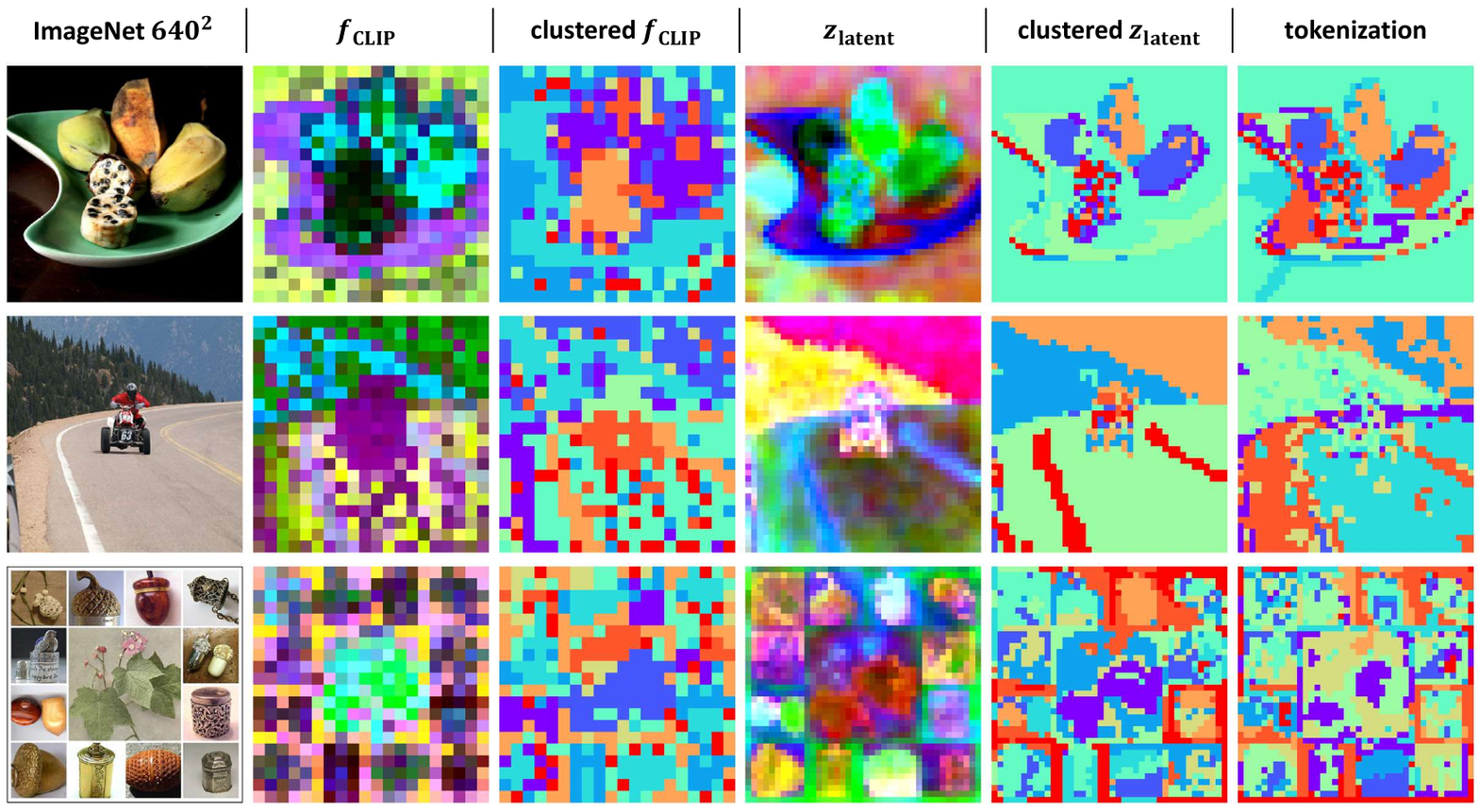}
    \caption{Feature Clustering versus Tokenization.  }
    \label{fig:meta}
\end{figure}

In order to study the behavior of PAT meta semantics and its association with feature clustering, we present the comparison of pretrained CLIP feature and corresponding the PAT latent feature in Fig\ref{fig:meta} for three representative scenarios: \textbf{obj}N/\textbf{bg}0, \textbf{obj}1/\textbf{bg}1 and N$\times$\textbf{obj}1/\textbf{bg}0 (N$\rightarrow$multiple, obj$\rightarrow$object, bg$\rightarrow$background). In general,
the CLIP feature clustering demonstrates the VLM capability of identifying major semantics, but its relatively low resolution representation for all background is noisy and prevent the discovery of smaller semantics. On the other hand, the segmentation guided features are clean and semantic separable for all scenarios. By comparing last two columns, the meta semantic shows the \textit{perception and semantic mixture} because PAT still possesses the composition pattern of the dominant objects and background pixel information, which supplements open vocabulary by visual (e.g., the road curb). 
To sum up, PAT trade-off the conflict learning branches for better performance.

\section{Discussion} \label{sec:discuss}

PAT explores learning visual tokens with certain level of semantic intuition, which is crucial for the VLM extensibility. We first build multi-resolution pretrained VLM feature tokenization for semantic-rich tokens which can bridge gap between image- and pixel-level understanding. With decoupled learning of perception and semantic compression, the contradictory encoding behavior between them is alleviated. The semantic-rich token representation thus enables parameter sharing of pixel and semantic decoding. 
Extensive experiments show that the PAT tokens is comprehensive and interpretable across feature pyramid revealing the cognition of visual concepts from low-level patterns to high-level abstractions, meanwhile improving the integrated segmentation model and achieving competitive segmentation OVSS performance. Furthermore, the tokenizer can be decoupled from PAT segmentation to serve other downstream tasks like generative tasks \cite{VAR} with more precise spatial-semantic controls.  

\bibliographystyle{IEEEbib}
\bibliography{c2vseg}

\end{document}